\documentclass[letterpaper]{article} 
\usepackage{aaai2026}  
\usepackage{times}  
\usepackage{helvet}  
\usepackage{courier}  
\usepackage[hyphens]{url}  
\usepackage{graphicx} 
\usepackage{natbib}  
\usepackage{caption} 
\usepackage{algorithm}
\usepackage{algorithmic}

\usepackage{amsthm}
\usepackage{amsmath}
\usepackage{multirow}
\theoremstyle{definition}
\newtheorem{definition}{Definition}

\usepackage{newfloat}
\usepackage{listings}
\DeclareCaptionStyle{ruled}{labelfont=normalfont,labelsep=colon,strut=off} 
\floatstyle{ruled}
\newfloat{listing}{tb}{lst}{}
\floatname{listing}{Listing}
\title{Lose the Order, Keep the Hierarchy: Deordering HTN Plans}
\author{
    Takudzwa Togarepi\textsuperscript{\rm 1},
    Gaspard Quenard\textsuperscript{\rm 1},
    Damien Pellier\textsuperscript{\rm 1},
    Humbert Fiorino\textsuperscript{\rm 1}
}
\affiliations{
    \textsuperscript{\rm 1}Université Grenoble Aples, France\\

    \{takudzwa.togarepi, gaspard.quenard\}@univ-grenoble-alpes.fr, \{damien.pellier, humbert.fiorino\}@imag.fr
}

\usepackage{bibentry}
\nocopyright 

\begin{document}

\maketitle

\begin{abstract}
Hierarchical Task Network (HTN) planning is a powerful planning formalism based on task decomposition. Although most of the literature studied plan generation, comparatively less attention has been paid to post-plan optimization. In particular, plan deordering has been extensively studied in classical planning but remains under-researched in the HTN setting. Plan deordering removes unnecessary ordering constraints between actions in a plan whilst keeping the plan valid. In this paper, we adapt two established plan deordering techniques from classical planning by extending the techniques to account for hierarchical decomposition constraints. We evaluate our proposed approaches on the IPC 2023 Partial-Order HTN benchmarks and we compare them against Optiplan, an HTN planner that generates partially ordered plans directly. Our results show a substantial reduction in number of ordering constraints in both our implementations. Although we also observe a reduction in critical path length, the improvements are less pronounced.
\end{abstract}


\section{Introduction}
Hierarchical Task Network (HTN) planning \citep{HtnNau} is planning by task decomposition; that is, recursively refining complex tasks until they become executable tasks. Among other planning formalisms, HTN planning is generally more expressive than STRIPS-style planning \citep{Erol_Htn_Expressiveness,  GEORGIEVSKI2015124}.

The main objective of an HTN planner is to determine how to refine an initial abstract (compound) task into a sequence of executable tasks. In simple terms, a plan is a sequence of primitive actions whose execution accomplishes a given task network. Existing HTN planners differ in the way they address an HTN problem. Some encode the given HTN problem into propositional logic in order to exploit existing specialized solvers \citep{Behnke_Chaos, Schreiber_Tree,Schreiber_Lilotane,Quenard_SibylSAT,quenard2025sibylsatopt}. Alternatively, other planners make use of heuristic search \citep{HollerBBB18,Holler_guiding_search,HollerBBB20}, whereas others translate HTN problems to a classical planning problem in order to exploit classical planners \citep{Alford16,Behnke22}.

While most HTN planning research has extensively studied plan generation, fewer works have addressed plan optimization \citep{ijcai2019p764,Behnke021}. In their recent survey, \citet{Bercher_Survey} define plan optimization as the task of improving a plan that has already been generated, and they also highlight several optimization techniques.

In practice, one important aspect of plan quality is plan-order flexibility. In some real-world applications of HTN planning, a totally ordered plan impose unnecessary ordering constraints at the expense of efficiency. For example, in a firefighting mission, a totally ordered plan could enforce that action $\mathit{dispatch}$-$\mathit{ground}$-$\mathit{crew}$ be executed before $\mathit{dispatch}$-$\mathit{helicopter}$. However, these two actions do not interfere and thus removing the ordering constraint between these two actions results in a flexible partial-order plan. Executing actions in parallel might improve execution speed and flexibility. This approach of post-plan optimization by removing ordering constraints on a plan while keeping it a valid plan is known as $\mathit{deordering}$ and has been extensively studied in the classical planning sense \citep{backstrom1998computational,Kambhampati_deordering, muise2011optimization,Bercher2020POPvsPOCL, SiddiquiH15_opt, MuiseBM16}. In particular, \citet{MuiseBM16} improved  deordering methods by using partially weighted MaxSAT which computes the minimal deorderings. A minimum deordering is a deordered plan in which one cannot remove any additional ordering constraints without invalidating the plan.

To the best of our knowledge, no prior work has explicitly studied plan deordering under HTN decomposition constraints. Although there is not much done in terms of HTN plan deordering, a few planners try to pursue a similar goal of having a partially ordered plan. One such planner is OptiPlan by \citet{FIRSOV_Thesis}, which produces a partially ordered plan directly. This motivates our interest in studying deordering in HTN planning. Our contributions in this paper are twofold. First, we adapt the PRF algorithm of \citet{backstrom1998computational} to the HTN setting, taking into account the hierarchical decomposition constraints. Second, we  compute minimal deorderings using a MaxSAT approach of \citet{muise2011optimization}, which we also adapt to the HTN setting.

We evaluate both implementations on the IPC 2023 Partial-Order HTN benchmarks. We apply the deordering techniques to plans generated by PANDA a partial-order planner \citep{Panda}. As a baseline, we compare the performance of the post-plan optimization against Optiplan. The results show a significant decrease in both the number of ordering constraints and the critical-path length of the plans in the post-plan optimization.

The remainder of this paper is structured as follows: following this introduction, we introduce the formalisms used throughout the paper. Afterward, we present the two implemented algorithms followed by a discussion of the results. Finally, we provide directions for future work and conclude the paper.

\section{Formalism}

We begin by introducing the formalisms for the techniques used throughout this paper

\subsection{STRIPS Planning Problem}

For the purpose of this paper, whenever we mention propositional planning, we restrict it to the STRIPS formalism by \citet{Bylander96}. In the STRIPS formalism, a planning task  is defined as a tuple  $\Pi$ = $\langle \mathcal{V, A, I, G}  \rangle$ where $\mathcal{V}$ refers to a set of propositional state variables, $\mathcal{A}$ is a finite set of actions, $\mathcal{I} \subseteq \mathcal{V}$ is the initial state and  $\mathcal{G} \subseteq \mathcal{V}$ is the goal condition. Each action $\mathit{a} \in \mathcal{A}$ is defined by a triple $\langle \mathit{pre(a)}, \mathit{add(a)}, \mathit{del(a)}\rangle$ where $\mathit{pre(a)} \subseteq \mathcal{V}$ is a set of preconditions, $\mathit{add(a)} \subseteq \mathcal{V}$ is a set of add effects, and $\mathit{del(a)} \subseteq \mathcal{V}$ is a set of delete effects. A state $\mathit{s} \subseteq \mathcal{V}$ is the set of state variables that are true. Under the closed-world assumption, every variable in $\mathcal{V}\backslash s$ is considered false. An action $\mathit{a}$ is applicable in $\mathit{s}$ if and only if $\mathit{pre(a)} \subseteq \mathit{s}$. The resulting successor state $\mathit{s}^\prime$ = ($\mathit{s} \backslash \mathit{del(a)}) \cup \mathit{add(a)}$. A plan (solution) $\pi$ = $\mathit{a}_1, \dots, \mathit{a}_n$, is a sequence of actions which when successively applied from the initial state lead to a goal state. A partially ordered plan (PO plan) maintains a partial-order among actions \citep{backstrom1998computational}.

\subsection{Hierarchical Task Network (HTN) Planning }
In this section, we present the HTN planning formalism by \citet{HtnDef} that we are going to use for the rest of the paper. HTN planning is centered on $\mathit{task}$ decomposition. A $\mathit{task}$ consists of a task name and an associated list of parameters. In HTN, tasks are classified into two categories: $\mathit{abstract}$ $\mathit{tasks}$ and $\mathit{primitive}$  $\mathit{tasks}$. A primitive task is analogous to an action in classical planning; that is, it can directly modify the state of the world. In contrast, abstract tasks cannot be executed directly and must be decomposed by applying decomposition $\mathit{methods}$.

A method $\mathit{m}$ is defined as a tuple ($ \mathit{c}, \mathit{tn}$) where $\mathit{c}$ is an abstract task and $\mathit{tn}$ a $\mathit{task}$ $\mathit{network}$. We refer to $\mathit{c}$ as the abstract task decomposed by method $\mathit{m}$, and to $\mathit{tn}$ as the $\mathit{task}$ $\mathit{network}$ that replaces $\mathit{c}$ when $\mathit{m}$ is applied. The resulting $\mathit{task}$ $\mathit{network}$ consists of  $\mathit{primitive}$ and/or $\mathit{abstract}$ $\mathit{tasks}$.

We now formally define a $\mathit{task}$ $\mathit{network}$ as follows:

\begin{definition}
(Task network). A task network $\mathit{tn}$ over a set of task names $\mathit{X}$ is a tuple ($\mathit{T}$, $\prec$, $\alpha$), where
\begin{itemize}
    \item $\mathit{T}$ is a finite set of tasks
    \item $\prec$ $\subseteq \mathit{T} \times \mathit{T}$ is a set of ordering constraints of the form ($\mathit{t_{i}<} \mathit{t_{j}}$) where $\mathit{t_{i}}, \mathit{t_{j}} \in$ $\mathit{T}$
    \item $\alpha$: $\mathit{T}$ $\rightarrow$ $\mathit{X}$ assigns a task name to each task in the network.
\end{itemize}
\end{definition}

$\mathit{TN_X}$ denotes a set of all task networks over $\mathit{X}$. A task network comprising only primitive tasks is known as a $\mathit{primitive}$ $\mathit{task}$ $\mathit{network}$. In their paper, \citet{HtnDef} defined an executable task network as one that contains a linearization of its tasks that is executable.

Next, we define a HTN planning problem.
\begin{definition}
    (Planning problem). A planning problem is a tuple $\mathcal{P}$ = ($\mathit{L,C,O,M,c_I,s_I,g}$), where
    \begin{itemize}
        \item $\mathit{L}$ is a finite set of propositions
        \item $\mathit{C}$ is a finite set of abstract tasks
        \item $\mathit{O}$ is a finite set of primitive tasks (actions)
        \item $\mathit{M} \subseteq \mathit{C \times}\mathit{TN_{\mathit{C} \cup \mathit{O}}  }$ a finite set of decomposition methods
        \item $\mathit{c_I} \in \mathit{C}$ initial abstract task
        \item $\mathit{s_I} \subseteq \mathit{L}$, the initial state
        \item $\mathit{g}$ the goal condition (possibly empty).
    \end{itemize}
\end{definition}

We interpret each primitive task (action) $\mathit{a} \in \mathit{O}$ as an action with preconditions and effects over $\mathit{L}$, i.e, $\mathit{pre(a)}$, $\mathit{add(a)}$, $\mathit{del(a)}$ $\subseteq \mathit{L}$. This allows us to properly reason about $\mathit{adders}$ and $\mathit{deleters}$. For proposition $\mathit{l} \in \mathit{L}$, let $\textbf{adders}$($\mathit{l}$) = $\{ \mathit{a} \in O \mid \mathit{l} \in \mathit{add}(\mathit{a}) \}$, and  $\textbf{deleters}$($\mathit{l}$) = $\{ \mathit{a} \in O \mid \mathit{l} \in \mathit{del}(\mathit{a}) \}$.

\begin{definition}
    (Decomposition). A decomposition refines an abstract task $\mathit{t}$ of a given task network $\mathit{tn}_1$ using a method $\mathit{m}$ = ($ \mathit{c}, \mathit{tn}_\mathit{m}$) with $\alpha$($\mathit{t}$) = $\mathit{c}$ resulting in a new task network $\mathit{tn}_2$. The decomposition replaces $\mathit{t}$ with $\mathit{tn}_\mathit{m}$. We write $\mathit{tn}_1$ $\rightarrow$ $\mathit{tn}_2$ for a single decomposition step and $\mathit{tn}_1$ $\rightarrow_D^{*}$ $\mathit{tn}_2$ for an arbitrary number of decompositions.
\end{definition}

We use a standard notion of decomposition trees of \citet{HtnDef}, however, we adopt a simplified formulation that is sufficient for our purposes.

\begin{definition}
    (Decomposition tree). Let $\mathit{D}_t$ be a decomposition tree that captures the refinement of an initial task network into a primitive task network. The root is the initial abstract task $\mathit{c_I}$ the primitive tasks are the leaves of the tree. The inner nodes represent the abstract tasks together with the methods.The tree induces a partial-order $\prec$ over its nodes.
\end{definition}
\begin{definition}
    (HTN solution). A task network $\mathit{tn_s}$ is a solution to an HTN planning problem $\mathcal{P}$ if and only if:
    \begin{enumerate}
        \item $\mathit{tn_s}$ is a primitive task network ,
        \item $\mathit{tn}$($\mathit{c_I}$) $\rightarrow_D^{*}$ $\mathit{tn_s}$,
        \item Any linearization of $\mathit{tn_s}$ is executable in $\mathit{s_I}$ and satisfies $\mathit{g}$.
    \end{enumerate}
\end{definition}

In this paper, we represent the HTN solution as $\langle \bar{a},\prec,\mathit{D}_t\rangle$ where $\bar{a} \subseteq \mathit{O}$ is a set of primitive tasks (actions); $\prec$ the ordering relations on these actions; $\mathit{D}_t$ represents the decomposition tree which captures the refinement of abstract tasks. Let, $\prec_\mathit{H}$ $ \subseteq$ $\mathit{D}_t$ denote the mandatory ordering constraints that are induced by the decomposition methods. Since the constraints are part of $\mathit{D}_t$ they should be preserved.


\section{HTN-PRF Algorithm}
In this section, we extend the classical PRF algorithm of \citet{backstrom1998computational} by integrating the mandatory constraints induced by the decomposition hierarchy. The PRF algorithm relies on $\mathit{simple}$ $\mathit{concurrency}$ to determine whether two distinct actions, $\mathit{a}_i$ and $\mathit{a}_j$ can execute in parallel. Their ordering can only be removed if they are not constrained by non-concurrency conditions.

\begin{definition}
    (Simple concurrency). Let $\mathit{a}_i$ and $\mathit{a}_j$ be two distinct actions. Actions $\mathit{a}_i$ and $\mathit{a}_j$ are considered non-concurrent, denoted by $\mathit{a}_i\#\mathit{a}_j$ if there exists an $\mathit{l} \in \mathit{L}$ such that:
    \begin{enumerate}
       \item $\mathit{l} \in \mathit{add}(\mathit{a}_i) \land \mathit{l} \in \mathit{pre}(\mathit{a}_j)$

       \item $\mathit{l} \in \mathit{add}(\mathit{a}_i) \land \mathit{l} \in \mathit{del}(\mathit{a}_j)$

       \item $\mathit{l} \in \mathit{pre}(\mathit{a}_i) \land \mathit{l} \in \mathit{del}(\mathit{a}_j)$
    \end{enumerate}
\end{definition}

If any of the above conditions holds for any $\mathit{l}$, then we preserve the ordering constraint between $\mathit{a}_i$ and $\mathit{a}_j$. In HTN planning, actions implicitly inherit the ordering constraints imposed by decomposition methods used during task refinement. These constraints should also therefore be conserved when deordering a plan.

\begin{algorithm}[tb]
\caption{HTN-PRF algorithm}
\label{alg:prf-hierarchy}
\textbf{Input}: Htn planning problem $\mathcal{P}$, HTN solution $\langle \bar{a},\prec,\mathit{D}_t\rangle$,  \\
\textbf{Output}: A valid PO-HTN solution
\begin{algorithmic}[1]

\FORALL{$\mathit{a}_i,\mathit{a}_j \in \bar{a}$ such that $\mathit{a}_i \prec \mathit{a}_j$}

   \IF{$(\mathit{a}_i,\mathit{a}_j)\in \prec_\mathit{H}$}
      \STATE Order $\mathit{a}_i \prec^\prime \mathit{a}_j$
    \ELSIF{$\mathit{a}_i \# \mathit{a}_j$}
        \STATE Order  $\mathit{a}_i$ $\prec^\prime$  $\mathit{a}_j$
    \ENDIF
\ENDFOR

\STATE \textbf{return} $\langle \bar{a},\prec^{\prime+},\mathit{D}_t\rangle$
\end{algorithmic}
\end{algorithm}

Algorithm~\ref{alg:prf-hierarchy} iterates over all ordering constraints in the primitive HTN solution. For each pair of ordered primitive actions $\mathit{a}_i \prec \mathit{a}_j$, the algorithm initially checks whether the ordering is induced by the decomposition tree. If so, the ordering is kept. Otherwise, the algorithm continues to perform the simple concurrency check. If the two actions are non-concurrent according to simple concurrency criterion, then we keep the ordering. Otherwise, the ordering between $\mathit{a}_i$ and $\mathit{a}_j$ is removed. The algorithm returns a partial-order HTN solution ($\langle \bar{a},\prec^{\prime+},\mathit{D}_t\rangle$) where $\bar{a}$ is a set of actions, $\prec^{\prime+} \subseteq \prec$ is a transitive closure of the ordering constraints and $\mathit{D}_t$ is the decomposition tree. Thus, the HTN-PRF algorithm only removes ordering constraints between $\mathit{a}_i$ and $\mathit{a}_j$ if the ordering is neither induced by the decomposition methods nor required by non-concurrency criteria.

Although the HTN-PRF deordering algorithm produces a partial-order plan, it does not guarantee an optimal deordering. To address this limitation, we propose an extension of the partial weighted MaxSAT algorithm of \citep{muise2011optimization}, which we refer to as HTN-MaxSAT deordering.

\section{HTN-MaxSAT deordering}

To generate optimal deorderings, we encode the deordering problem as a partial weighted MaxSAT problem. Given an HTN planning problem $\mathcal{P}$ = ($\mathit{L,C,O,M,c_I,s_I,g}$), let:
\begin{itemize}
    \item $\bar{a} \subseteq \mathit{O}$ be the primitive tasks (actions) of an HTN plan,
    \item $\mathit{a}_I$ be an initial dummy action,
    \item $\mathit{a}_G$ be a dummy goal action.
\end{itemize}

For every pair of actions $\mathit{a}_i$, $\mathit{a}_j$ $\in$ $\bar{a}$, we represent an ordering constraint between them using a variable $\kappa(\mathit{a}_i,\mathit{a}_j)$. The encodings of \citet{muise2011optimization} used an action-membership variable to represent an action present in the partial-order plan. However, in our setting, this is not necessary as every action in $\bar{a}$ must remain present in the final partial-order plan after deordering. Since all actions in $\bar{a}$ are part of a valid HTN solution generated through the decomposition hierarchy, they must be preserved to maintain correctness.

Partial weighted MaxSAT allows us to differentiate between two types of clauses, namely, hard clauses and soft clauses. The objective of partially weighted MaxSAT is to satisfy all hard clauses and maximize the total weight associated with the soft clauses. We begin by presenting the hard clauses of our encoding.

Let $\prec_\mathit{H}$ be the set of mandatory ordering constraints induced by the decomposition methods. We must preserve all ordering constraints in $\prec_\mathit{H}$:
\begin{equation}
    \kappa(\mathit{a}_i,\mathit{a}_j), \quad  \forall (\mathit{a}_i,\mathit{a}_j) \in \prec_\mathit{H}
    \label{eq:hierarchy}
\end{equation}
Furthermore, we enforce additional structural properties of a valid partially ordered plan:

\begin{equation}
    (\neg \kappa(\mathit{a}_i,\mathit{a}_i)) \\
    \label{eq:noselfloop}
\end{equation}
\begin{equation}
    \kappa(\mathit{a}_I,\mathit{a}_i) \land \kappa(\mathit{a}_i,\mathit{a}_G) \ \\
    \label{eq:init_goal}
\end{equation}
\begin{equation}
    \kappa(\mathit{a}_i,\mathit{a}_j) \land \kappa(\mathit{a}_j,\mathit{a}_k) \to \kappa(\mathit{a}_i,\mathit{a}_k) \\
    \label{eq:transitive}
\end{equation}

(\ref{eq:noselfloop}) forbids self loops; (\ref{eq:init_goal}) ensures that each action is between $\mathit{a}_I$ and $\mathit{a}_G$, we also assume $\mathit{a}_I \neq \mathit{a}_i \neq \mathit{a}_G$; (\ref{eq:transitive}) captures the transitive relations of the ordering constraints; the combination of (\ref{eq:noselfloop}) and (\ref{eq:transitive}) ensures our partial-order plan will be acyclic. For causal correctness, we define the following:

\begin{equation}
     \gamma(\mathit{a}_i, \mathit{a}_j, \mathit{l}) \equiv \bigwedge\limits_{\mathit{a}_k \in \text{deleters}(\mathit{l})} (\kappa(\mathit{a}_k,\mathit{a}_i) \lor \kappa(\mathit{a}_j,\mathit{a}_k))\\
    \label{eq:preconditions}
\end{equation}
which forbids any action $\mathit{a}_k$ that deletes $\mathit{l}$ from occurring between $\mathit{a}_i$ and $\mathit{a}_j$ when $\mathit{a}_i$ produces $\mathit{l}$ for $\mathit{a}_j$. Additionally, every precondition should be supported:
\begin{equation}
    \bigwedge\limits_{\mathit{l}\in \mathit{pre}(\mathit{a}_j)} \bigvee\limits_{\mathit{a}_i \in \text{adders}(\mathit{l})} (\kappa(\mathit{a}_i,\mathit{a}_j) \land \gamma(\mathit{a}_i, \mathit{a}_j, \mathit{l}))
    \label{eq:achivers}
\end{equation}

A combination of (\ref{eq:preconditions}) and (\ref{eq:achivers}) enforces the condition that for each precondition $\mathit{l}$ of $\mathit{a}_j$, there exists an action $\mathit{a}_i$ which adds $\mathit{l}$ and precedes $\mathit{a}_j$, while forbidding any action $\mathit{a}_k$ that deletes $\mathit{l}$ from occurring between $\mathit{a}_i$ and $\mathit{a}_j$.

To conclude the hard clauses, we ensure we only remove unnecessary ordering constraints without reversing the original ordering in our sequential plan. Assuming our sequential plan was originally ordered as $\mathit{a}_0, \dots ,\mathit{a}_k$ then,:

\begin{equation}
    \neg \kappa(\mathit{a}_j,\mathit{a}_i), \quad 0 \leq \ i < j \leq k
    \label{eq:ordering}
\end{equation}

For optimization, we add a soft unit clause for each ordering constraint that neither contains $\mathit{a}_I$ nor $\mathit{a}_G$ and is not induced by the hierarchy:

\begin{equation}
    w(\neg \kappa(\mathit{a}_i,\mathit{a}_j)) = 1
    \label{eq:soft-clause}
\end{equation}

\appendix

\section{Evaluation}
We conducted the experiments on a Linux laptop with an Intel(R) Core(TM) Ultra 9 185H processor and 32GB of RAM. Each problem instance was limited to a runtime of 30 minutes and 10GB of  memory. We evaluate the effectiveness of our implementations using two metrics: (i) the number of ordering constraints, (ii) the length of the critical path. In this paper, we consider the critical path length as the longest chain of sequential actions. We compare our resulting partial-order plans to those produced by Optiplan \citep{FIRSOV_Thesis}. For our implementations\footnote{https://github.com/gaspard-quenard/sibylsat/tree/deordering}, we also report the percentage improvements relative to the original plans denoted as \textbf{\%Diff} in our tables.

We generate sequential plans using PANDA \citep{Panda} which are then used as input to our HTN-PRF and HTN-MaxSAT implementations. We conduct our experiments using the IPC 2023 Partial-Order HTN benchmarks. When we compare Optiplan to our implementations we only report domains and problem instances that are solved by all the approaches.

The values shown in the tables are the mean value for each of the domains. The percentage improvement (\textbf{\%Diff}) for both the number of ordering constraints and critical path length are independently computed per domain relative to the original sequential plan. Given an init value (\textbf{Init}) and a deordered value (\textbf{Final}), we calculate $(\textbf{Init}-\textbf{Final})/\textbf{Init} \times 100$.
The last row of the tables contain weighted values over all problem instances computed by weighing the mean by the number of instances.

To enhance readability we adopt a short-hand notation for our implementations, H$_P$ refers to HTN-PRF, whereas H$_M$ and OP refer to HTN-MaxSAT and Optiplan respectively.

\begin{table}[!h]
\centering
\small
\setlength{\tabcolsep}{1.5pt}
\begin{tabular}{l||ccc||ccc}
\hline
\multirow{2}{*}{\textbf{Domain}}
& \multicolumn{3}{c||}{\textbf{mean constraints}}
& \multicolumn{3}{c}{\textbf{mean critical path}} \\
\cline{2-7}
& \textbf{H$_P$} & \textbf{H$_M$} & \textbf{OP}
& \textbf{H$_P$} & \textbf{H$_M$} & \textbf{OP} \\
\hline
PCP (9)        & \textbf{240.77} & \textbf{240.77} & 243.56 & \textbf{21.11} & \textbf{21.11} & 22.33 \\
Rover (4)      & 32 & \textbf{28.50} & 522.75 & 7.00 & \textbf{6.70} & 25.75 \\
Satellite (25) & 46.12 & 46.12 & \textbf{30.32} & 8.72 & 8.72 & \textbf{6.60} \\
Transport (11) & 188.64 & 188.64 & \textbf{155.36} & 16.55 & 16.55 & \textbf{12.64} \\
Um-Translog (22) & 138.55 & 138.55 & \textbf{138.22} & \textbf{15.00} & \textbf{15.00} & 15.81 \\
\hline
ALL (71)       & 120.72 & \textbf{120.52} & 137.90 & 13.35 & \textbf{13.34} & 13.46 \\
\hline
\end{tabular}
\caption{Mean ordering constraints and mean critical path of the partial-order plans. Numbers next to the domain names are the sum of problem instances evaluated for all methods}
\label{tab:constraints-summary}
\end{table}

Table~\ref{tab:constraints-summary} reports the mean number of ordering constraints together with the mean critical path length of our two HTN approaches and Optiplan. In comparison, HTN-MaxSAT produces a plan with marginally lower mean number of ordering constraints to that of HTN-PRF. However, the difference is more pronounced when compared to Optiplan. This is despite the fact that Optiplan directly outputs a partial-order plan and the HTN approaches use a sequential plan generated by PANDA. The higher number of ordering constraints observed in Optiplan is due to Optiplan generating larger plans than those by PANDA. Across all approaches, the mean critical path length is comparable, with HTN-MaxSAT producing a partial-order plan with a slightly shorter mean critical path.

\begin{table}[!h]
\centering
\small
\setlength{\tabcolsep}{1.5pt}
\begin{tabular}{l||ccc||ccc}
\hline
\multirow{2}{*}{\textbf{Domain}}
& \multicolumn{3}{c||}{\textbf{H$_P$}}
& \multicolumn{3}{c}{\textbf{H$_M$}} \\
\cline{2-7}
& \textbf{Init} & \textbf{Final} & \textbf{\%Diff}
& \textbf{Init} & \textbf{Final} & \textbf{\%Diff} \\
\hline
PCP (9)        & 240.77 & 240.77 & \textbf{0} & 240.77 & 240.77 & \textbf{0} \\
Rover (4)      & 39 & 32 & 17.95 & 39 & 28.50 & \textbf{26.92} \\
Satellite (25) & 63.76 & 46.12 & \textbf{27.67} & 63.76 & 46.12 & \textbf{27.67} \\
Transport (11) & 214.73 & 188.64 & \textbf{12.15} & 214.73 & 188.64 & \textbf{12.15} \\
Um-Translog (22) & 146.73 & 138.55 & \textbf{5.57} & 146.73 & 138.55 & \textbf{5.57} \\
\hline
ALL (71)       & 133.90 & 120.72 & 9.84 & 133.90 & 120.52 & \textbf{9.99} \\
\hline
\end{tabular}
\caption{Ordering constraints percentage improvements}
\label{tab:deordering-summary}
\end{table}

Table \ref{tab:deordering-summary} shows that apart from the PCP domain, both HTN-PRF and HTN-MaxSAT achieve significant reductions in the number of ordering constraints in the partial-order plan. In particular, the largest improvement is observed in the Satellite domain using HTN-MaxSAT algorithm with a mean reduction of 27.67\%. HTN-MaxSAT performed at least as well as HTN-PRF in all domains and notably outperforms it in the Rover domain by close to 9\%. In domains such as Transport, we observe a correlation between number of trucks and number of ordering constraints removed. However, PANDA did not find solutions for most of the harder problems in Transport, which limits the observed improvements.

\begin{table}[!h]
\centering
\small
\setlength{\tabcolsep}{1.5pt}
\begin{tabular}{l||ccc||ccc}
\hline
\multirow{2}{*}{\textbf{Domain}}
& \multicolumn{3}{c||}{\textbf{H$_P$}}
& \multicolumn{3}{c}{\textbf{H$_M$}} \\
\cline{2-7}
& \textbf{Init} & \textbf{Final} & \textbf{\%Diff}
& \textbf{Init} & \textbf{Final} & \textbf{\%Diff} \\
\hline
PCP (9)        & 21.11 & 21.11 & \textbf{0} & 21.11 & 21.11 & \textbf{0} \\
Rover (4)      & 9.25 & 7.00 & 24.32 & 9.25 & 6.70 & \textbf{27.57} \\
Satellite (25) & 10.96 & 8.72 & \textbf{20.44} & 10.96 & 8.72 & \textbf{20.44} \\
Transport (11) & 20 & 16.55 & \textbf{17.25} & 20 & 16.55 & \textbf{17.25}  \\
Um-Translog (22) & 15.60 & 15.00 & \textbf{3.85} & 15.60 & 15.00 & \textbf{3.85} \\
\hline
ALL (71)       & 14.99 & 13.35 & 10.91 & 14.99 & 13.34 & \textbf{11.03} \\
\hline
\end{tabular}
\caption{Critical path percentage improvements}
\label{tab:critical-path-summary}
\end{table}

The reduction in critical path length in Table~\ref{tab:critical-path-summary} follow a similar pattern to the reductions in ordering constraints in Table~\ref{tab:deordering-summary}. The largest percentage reduction is observed in the Rover domain. In contrast, the PCP domain remains totally ordered, with no reduction in both ordering constraints or critical path length.



\section{Conclusion}
In this paper, we proposed two approaches for HTN plan deordering. The first, HTN-PRF, preserves ordering constraints enforced by the hierarchy or required by simple concurrency and removes all others. The second, HTN-MaxSAT, encodes the problem as a partially weighted MaxSAT formulation and guarantees a minimal deordering for a given input plan.

Our results show that both approaches significantly reduce ordering constraints and the critical path length, resulting in more flexible partial-order plans. However, their effectiveness depends heavily on the quality of the input sequential plan. Although larger reductions were observed on harder instances, most harder instances were unsolved.
\section{Acknowledgments}
This work has been supported by the French National Research Agency (ANR) under the project ANR-24-CE10-0893 (Prod4Human).

\end{document}